\documentclass[lettersize,journal]{IEEEtran}
\usepackage{amsmath,amsfonts}
\usepackage[caption=false,font=normalsize,labelfont=sf,textfont=sf]{subfig}
\usepackage{graphicx}
\usepackage{cite}
\usepackage{booktabs}
\usepackage{multirow}
\usepackage{siunitx}
\usepackage{threeparttable}
\usepackage{xcolor}
\usepackage{pifont}
\usepackage{marvosym}
\usepackage{xspace}
\usepackage{soul}
\definecolor{circlegreen}{HTML}{00A651}
\definecolor{circlered}{HTML}{D62728}
\definecolor{fillgreen}{HTML}{009E73}
\definecolor{fillblue}{HTML}{0072B2}
\definecolor{fillred}{HTML}{D55E00}

\newcommand{\cmark}{\textcolor{green!65!black}{\ding{51}}}
\newcommand{\xmark}{\textcolor{red}{\ding{55}}}
\newcommand{\figref}[1]{Fig.~\ref{#1}}
\newcommand{\ie}{\textit{i.e.}\xspace}
\begin{document}

\title{CARE: Anti-entanglement Ultrasound Image Segmentation via Channel-Aware Region Extrication}

\author{Weixin Xu, Yuting Lu, Luqi Gong, Qing Guo, Ziliang Wang\textsuperscript{\Letter}, and Yun Xing\textsuperscript{\Letter}
\thanks{(Corresponding authors: Ziliang Wang (wangziliang@pku.edu.cn) and Yun Xing (yxing.wiwin@gmail.com).)
}}

\maketitle

\begin{abstract}
Accurate ultrasound image segmentation is fundamentally challenged by target-context entanglement, where lesion cues are easily mixed with surrounding tissues and artifacts of similar appearance. Although existing methods often localize suspicious regions reasonably well, they remain vulnerable to ambiguous predictions because they mainly strengthen feature extraction or context aggregation, rather than explicitly organizing how lesion and interference cues are represented and distinguished. To address this limitation, we propose Channel-Aware Region Extrication (CARE), a segmentation framework that improves ultrasound segmentation by progressively extricating lesion evidence from visually entangled context. Instead of merely reweighting features, CARE explicitly separates encoded responses according to their lesion relevance and then re-evaluates the resulting complementary representations through reciprocal region interaction, so that suppressed lesion cues can be recovered while misleading contextual activations are corrected. In this way, CARE promotes target-context discrimination directly in the learned representation, without sacrificing localization quality. Extensive experiments on BUSI, BUSIS, and TN3K benchmarks show that CARE consistently achieves superior performance, thereby validating representation extrication as an effective solution for addressing the inherent visual ambiguity in ultrasound segmentation.
\end{abstract}

\begin{IEEEkeywords}
Image Segmentation, Ultrasound Image, Disentanglement.
\end{IEEEkeywords}

\section{Introduction}
\begin{figure*}
    \centering
  \includegraphics[width=\linewidth]{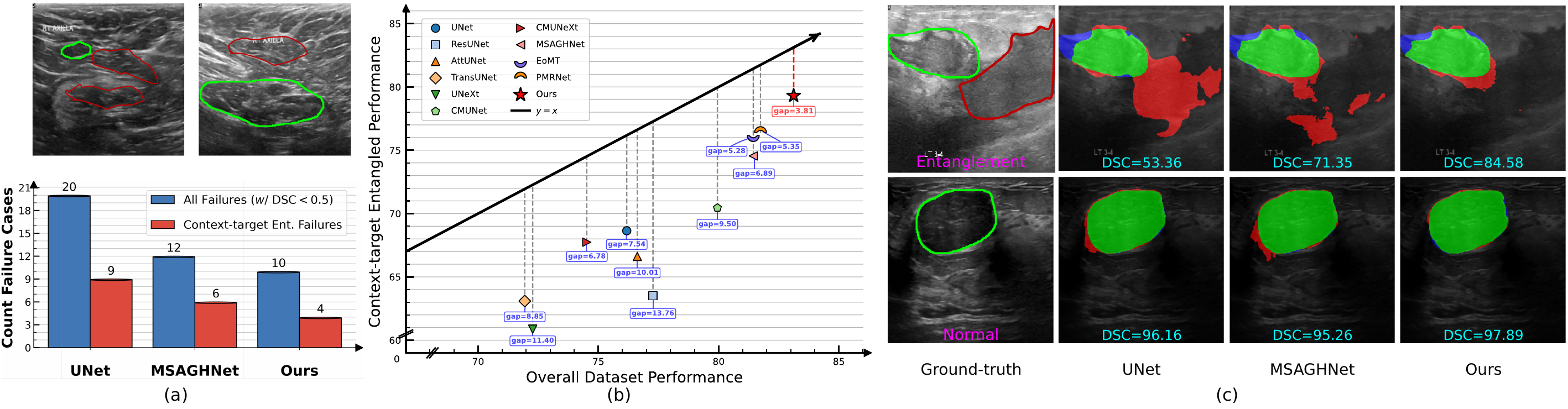}
    \vspace{-0.8cm}

    \caption{(a) Representative target-context entangled samples from BUSI dataset (top) and their proportion among model failures on the test set (bottom). Lesions and entangled contexts are outlined in \textcolor{circlegreen}{green} and \textcolor{circlered}{red} respectively. (b) DSC performance of existing ultrasound segmentation models on the full BUSI test set (x-axis) and the 46 entangled samples identified by our user study (y-axis). ``gap'' highlights the performance degradation posed by the entanglement challenge. (c) Segmentation results of UNet, MSAGHNet and our CARE for a pair of ``Entanglement \emph{vs.} Normal'' ultrasound images. \textcolor{fillgreen}{Green}, \textcolor{fillblue}{blue}, and \textcolor{fillred}{red} denote correct, missed and erroneous predictions respectively.}
    \label{fig:main}
    \vspace{-0.6cm}
\end{figure*}

\IEEEPARstart{M}{edical} image segmentation is a fundamental task in multimedia-based medical image analysis, as it requires dense visual understanding of anatomical structures and lesions for downstream clinical applications~\cite{yao2024cnn}.
Among different imaging modalities, ultrasound is widely used in breast and thyroid examination because it is real-time, low-cost, and radiation-free~\cite{iacob2024evaluating}.
Yet, compared with other modalities, ultrasound poses a distinctive challenge for segmentation: the target is often embedded in complex visual context with weak contrast and strong ambiguity~\cite{xue2021global, wang2021residual}.
Owing to acoustic scattering and related imaging artifacts, lesion regions may share highly similar appearance with surrounding tissues in texture, intensity, and local geometry.
As a result, ultrasound segmentation depends not only on accurate localization, but also on learning discriminative representations that can separate subtle lesion patterns from visually entangled background responses.

Existing methods mainly improve ultrasound segmentation through stronger feature extraction or context modeling~\cite{ronneberger2015u, zhang2018road, oktay2018attention, chen2021transunet},
yet they insufficiently address a more fundamental challenge: the visual entanglement between lesion cues and background regions of similar appearance~\cite{wu2023cross}.
In practice, many models can already localize suspicious regions coarsely, but still struggle to distinguish true lesions from confusing structures, leading to boundary leakage, over-segmentation, or missing subtle lesion parts~\cite{xue2021global, wang2021residual}.
This limitation is especially pronounced in ultrasound, where localization alone is often insufficient once lesion evidence is mixed with visually similar surrounding responses in the learned representation~\cite{maity2015comparative, gu2020context}.
Consequently, the bottleneck is not merely how to aggregate more context, but how to organize lesion and background cues into representations that remain discriminative under severe visual ambiguity.

As illustrated in \figref{fig:main}, this challenge is both common and consequential.
Representative examples in \figref{fig:main}(a) show that lesions are often embedded in visually similar surrounding tissues or artifacts, while the results in \figref{fig:main}(b) indicate a clear performance degradation on such entangled cases relative to the overall test set.
The qualitative comparison in \figref{fig:main}(c) further suggests that current models often fail not because they cannot roughly localize the lesion, but because they cannot form sufficiently discriminative representations to suppress misleading context responses.
Taken together, these observations point to a limitation in current representation learning: lesion and interference cues are often mixed rather than selectively organized for reliable dense prediction under visual ambiguity.

To address this limitation, we propose Channel-Aware Region Extrication (CARE), a segmentation framework that improves ultrasound segmentation by progressively extricating lesion evidence from visually entangled context.
Rather than relying solely on stronger feature aggregation, CARE promotes target-context discrimination directly in the learned representation.
In essence, CARE treats ultrasound segmentation as a process of representation extrication, where lesion evidence is progressively separated from, and re-evaluated against, confusing context.
It organizes encoded responses according to their lesion relevance and further refines their interactions so that suppressed lesion cues can be recovered while misleading contextual activations are corrected.
Under this formulation, Channel-wise Relevance Decoupling (CRD) exposes the latent separation between lesion-relevant and interference-dominant responses, and Mutual Region Querying (MRQ) further consolidates this separation through reciprocal context-aware refinement.
In this way, CARE reduces representation ambiguity while preserving the localization capability required for accurate lesion delineation.
Different from prior methods that mainly enhance feature extraction or context aggregation, CARE explicitly targets how lesion and background cues are separated and refined in the learned representation.

In summary, this study contributes to the ultrasound image segmentation research with following aspects:
\begin{itemize}
    \item We propose Channel-Aware Region Extrication (CARE), a segmentation framework for ultrasound images that addresses target-context entanglement, where lesions are easily confused with visually similar surrounding background, such as tissues and artifacts.

    \item We introduce a disentanglement-and-refinement mechanism for robust target-context discrimination. By decoupling target-relevant and interference-related representations and enabling reciprocal cross-branch correction, CARE reduces representation ambiguity and improves lesion localization and delineation.

    \item We conduct extensive experiments on three public ultrasound segmentation benchmarks, namely BUSI, BUSIS, and TN3K. The results show that CARE consistently outperforms existing methods, demonstrating strong effectiveness and robustness for ultrasound image segmentation under severe visual ambiguity.
\end{itemize}

\section{Related Work}
\label{sec:relate}

\textbf{Semantic Image Segmentation.}
Semantic segmentation has evolved from fully convolutional architectures~\cite{long2015fully} to Transformer-based \cite{xie2021segformer, wang2021pyramid} and hybrid CNN-Transformer models \cite{hatamizadeh2022unetr}. CNN-based methods improve dense prediction through hierarchical feature extraction, skip connections, and multi-scale context aggregation, as exemplified by Deeplab~\cite{chen2017deeplab}, PSPNet~\cite{zhao2017pyramid}, and related encoder--decoder variants. Transformer-based models further strengthen long-range dependency modeling via self-attention, with representative examples including ViT~\cite{dosovitskiy2020image}, SETR~\cite{zheng2021rethinking}, PVT~\cite{wang2021pyramid}, and SegFormer~\cite{xie2021segformer}. Hybrid architectures, such as TransUNet~\cite{chen2021transunet} and UNETR~\cite{hatamizadeh2022unetr}, further combine local texture modeling and global context reasoning within unified encoder--decoder frameworks. Despite their strong performance, these methods mainly emphasize feature extraction and context fusion. For ultrasound images, however, the major challenge is not only capturing richer context, but also distinguishing lesions from visually similar background regions and artifacts. Since existing segmentation paradigms do not explicitly model such target-context entanglement, their discrimination ability remains limited under severe visual ambiguity.

\textbf{Medical Image Segmentation.}
Medical image segmentation requires dense pixel-wise classification of anatomical structures and lesions, and has been dominated by encoder-decoder architectures following UNet~\cite{ronneberger2015u}. Subsequent CNN-based variants improve upon UNet along several axes: ResUNet~\cite{zhang2018road} and DenseUNet~\cite{zhou2022denseunet} incorporate residual and dense connections for stronger feature reuse; U-Net++~\cite{zhou2019unet++} introduces nested skip connections for richer multi-scale feature fusion; and Attention UNet~\cite{oktay2018attention} applies gated attention to suppress irrelevant background regions. Despite these advances, CNN-based methods are fundamentally constrained by the local receptive field of convolution, limiting their capacity to model long-range structural dependencies. This limitation is particularly pronounced in ultrasound image segmentation, where targets exhibit high shape variability and low contrast boundaries that require broad contextual understanding to delineate reliably.
To overcome this constraint, hybrid and pure-transformer architectures have been explored. TFCNs~\cite{li2022tfcns} and CASTformer~\cite{chenyu2022class} combine convolutional encoders with transformer modules to jointly capture local texture and global context. Swin-UNet~\cite{cao2022swin} adopts a fully transformer-based encoder-decoder using the Swin Transformer~\cite{liu2021swin} for hierarchical feature extraction. While these methods demonstrate improved global modeling, they are designed primarily on MRI and CT modalities, where structures are well-defined with high contrast. Ultrasound images, by contrast, are characterized by speckle noise, acoustic shadowing, and ambiguous target boundaries, which existing paradigms do not explicitly address. This motivates the need for segmentation models tailored to the unique properties of ultrasound imaging.

\section{Methodology}
\label{sec:method}

Due to the reliance on acoustic scattering during image formation, ultrasound images suffer from a more pronounced context-target entanglement compared to other medical modalities (See \figref{fig:main} (a)).

In this work, we posit that \textit{\ul{the inferior performance of existing ultrasonic segmentation methods stems from the failure of distinguishing the target regions from its image background}}.
Therefore,
we first \ding{182} provide a fundamental formulation of the context-target entanglement challenge specific to ultrasound segmentation task.
Then, we \ding{183} verify the influences and investigate the culprit of context-target entanglement issue over existing models by conducting extensive empirical studies with representative ultrasonic segmentation methods.
Based on the insights derived from the empirical analysis, we \ding{184} propose our novel solution, \ie, Channel-Aware Region Extrication (CARE), for discriminating the target from given ultrasound images where the background is crowded with appearance deceits.

\subsection{Problem Statement}

Given a medical ultrasonic image $\mathbf{I}\in\mathbb{R}^{C\times H\times W}$, the segmentation task aims to predict a binary mask $\mathbf{M}$ as
\begin{equation}
    \mathbf{M} = S_\phi(\mathbf{I}),
    \label{eq:segmentation}
\end{equation}
where $S_\phi(\cdot)$ denotes a segmentation model parametrized by $\phi$, and mask $\mathbf{M} \in \{0,1\}^{H\times W}$ delineates the pixel regions where lesion resides.
Fundamentally, ultrasound imaging relies on acoustic scattering to capture the pathological information.
As normal and lesion regions exhibit minimal contrast in their acoustic properties, the resulting images typically contain regions with intensity distributions that highly similar to the target region (See \figref{fig:main} (a)).

To fully uncover such intrinsic characteristics of ultrasound imaging, we formulate the ultrasonic image $\mathbf{I}$ as a composition
\begin{equation}
    \mathbf{I} = \mathcal{B} \cup \mathcal{O}_t \cup \{\mathcal{O}_d^i\}_{i=1}^n,
    \label{eq:region_decomposition}
\end{equation}
where $\mathcal{O}_t$ denotes the target region of interest.
$\mathcal{B}$ represents the image background and $\{\mathcal{O}_d^i\}_{i=1}^n$ comprises distracting regions, namely the context, that exhibit similar intensity patterns to the target.
Moreover, we note that the spatial relationships between any two regions $\mathcal{O}_i$ and $\mathcal{O}_j$ satisfy either $\mathcal{O}_i \cap \mathcal{O}_j = \emptyset$ (disconnected/disjoint) or $\mathcal{O}_i \cap \mathcal{O}_j \neq \emptyset$ (connected/overlapping).
Essentially, the context-target entanglement challenge arises when $\exists ~\mathcal{O}_d \in \{\mathcal{O}_d^i\}$ such that
\begin{equation}
    s(\mathcal{O}_d, \mathcal{O}_t) < \epsilon, \quad \textit{s.j.} \quad \mathcal{O}_d \cap \mathcal{O}_t = \emptyset
    \label{eq:entanglement}
\end{equation}
where $s(\cdot,\cdot)$ measures the appearance similarity, and $\epsilon$ defines the maximum threshold for distinguishing benign and malignant regions.

Accordingly, our goal with ultrasound image segmentation is to design the model $S_\phi(\cdot)$ that accurately identifies $\mathcal{O}_t$ by discriminating it from distracting regions $\{\mathcal{O}_d^i\}$ whose appearance similarity $s(\mathcal{O}_d, \mathcal{O}_t)$ approaches the threshold $\epsilon$, while preserving sensitivity to the spatial disjointness condition $\mathcal{O}_d \cap \mathcal{O}_t = \emptyset$ that distinguishes true distractors from infiltration zone.
It is worth to note that Eq.~\eqref{eq:entanglement} does not account for multi-lesion scenarios.
Nevertheless, we refer readers to the supplemental material where we discuss and demonstrate that our designing is also capable of handling multi-lesion situations.

\begin{figure*}[ht]
    \centering
    \includegraphics[width=\textwidth]{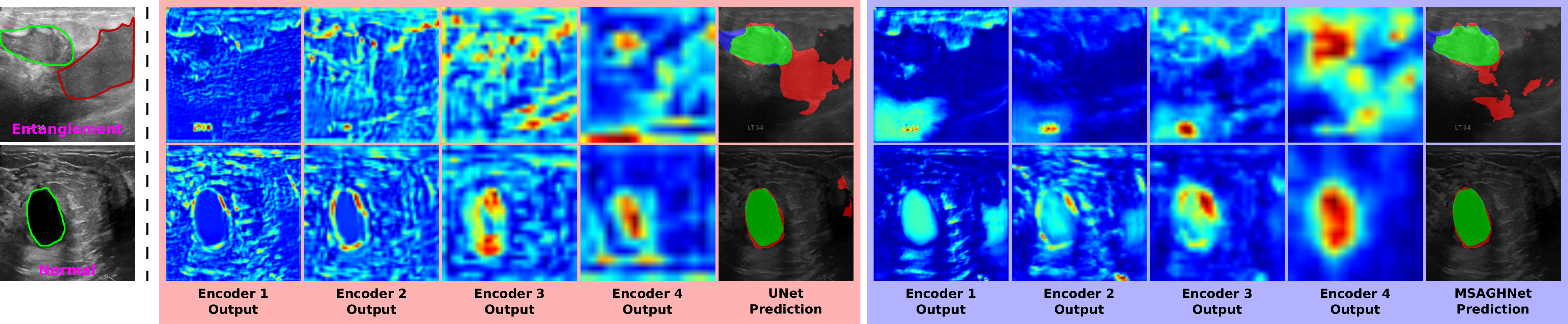}
    \vspace{-0.6cm}
    \caption{Visualization of the feature maps for a typical pair of ``Entanglement \emph{v.s.} Normal'' ultrasound images during the encoding process, \ie, Encoder 1-4, of UNet and MSAGHNet. Note that the more intensive the red color, the more attention the model focuses on the corresponding regions, and \emph{vice versa} for blue colors.}
    \label{fig:emp_a}
    \vspace{-0.6cm}
\end{figure*}

\subsection{Empirical Motivation}
\begin{figure}[t]
    \centering
    \includegraphics[width=0.9\linewidth]{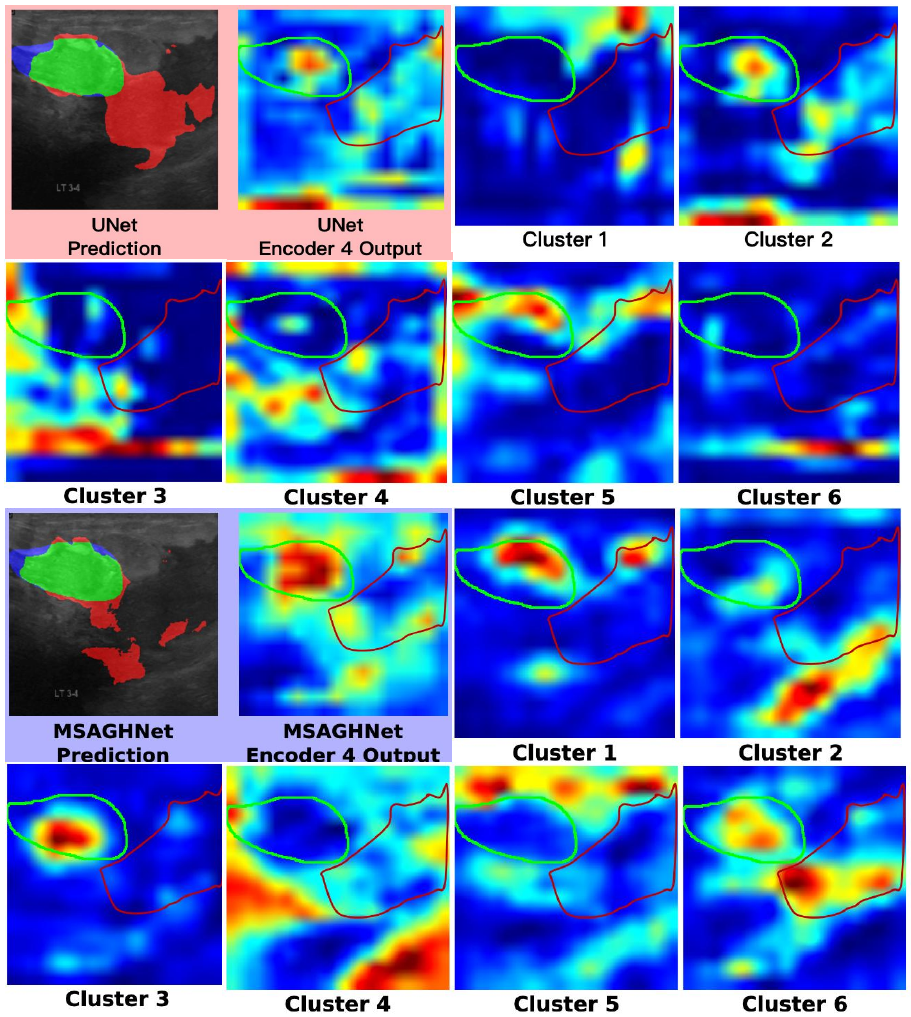}
    \vspace{-0.4cm}
    \caption{Illustration of the clustering results over channel-wise features (encoder 4) for the ``Entanglement'' case in (\figref{fig:emp_a}). The delineations of the target and entanglement regions are mapped to the clustering results for a clear comparison. For the prediction results, \textcolor{fillgreen}{green}, \textcolor{fillblue}{blue}, and \textcolor{fillred}{red} denote correct, missed and erroneous regions respectively. In heatmaps, lesions and entangled contexts are outlined in \textcolor{circlegreen}{green} and \textcolor{circlered}{red} respectively.}
    \label{fig:emp_b}
    \vspace{-0.8cm}
\end{figure}

To verify the practical impact of target-context entanglement, we analyze two representative ultrasound segmentation models, namely UNet and MSAGHNet, on the BUSI dataset.
A user study identifies 46 entangled samples out of 129 test images.
We find that these samples account for a substantial portion of model failures, indicating that target-context entanglement is not an occasional phenomenon but a common source of segmentation errors.
Moreover, when evaluated separately on the entangled subset, both models show clear performance degradation relative to their overall test set results.
Specifically, MSAGHNet suffers a DSC gap of 6.89\%, while UNet exhibits an even larger gap of 7.54\%, as shown in \figref{fig:main}(a)(b).
These results suggest that target-context entanglement is a critical factor limiting existing ultrasound segmentation methods.

To further understand this limitation, we inspect intermediate encoder responses on representative entangled cases.
As illustrated in \figref{fig:main}(c) and \figref{fig:emp_a}, existing models can often coarsely localize lesion regions, but fail to reliably suppress visually similar distractors.
This ambiguity appears early in the encoding process, suggesting that the bottleneck lies less in coarse localization and more in insufficient target-context discrimination in the learned representation.
In other words, lesion-relevant and interference-related cues are often mixed rather than selectively organized during feature encoding.

We further inspect channel-wise responses in the last encoder.
As illustrated in \figref{fig:emp_b}, lesion-related and distracting responses exhibit partial separation across channels, yet existing models do not effectively organize or exploit this distinction during feature encoding, thereby demonstrating the motivation for our proposed channel-aware design.

\begin{figure*}[ht]
   \centering
   \includegraphics[width=0.9\textwidth]{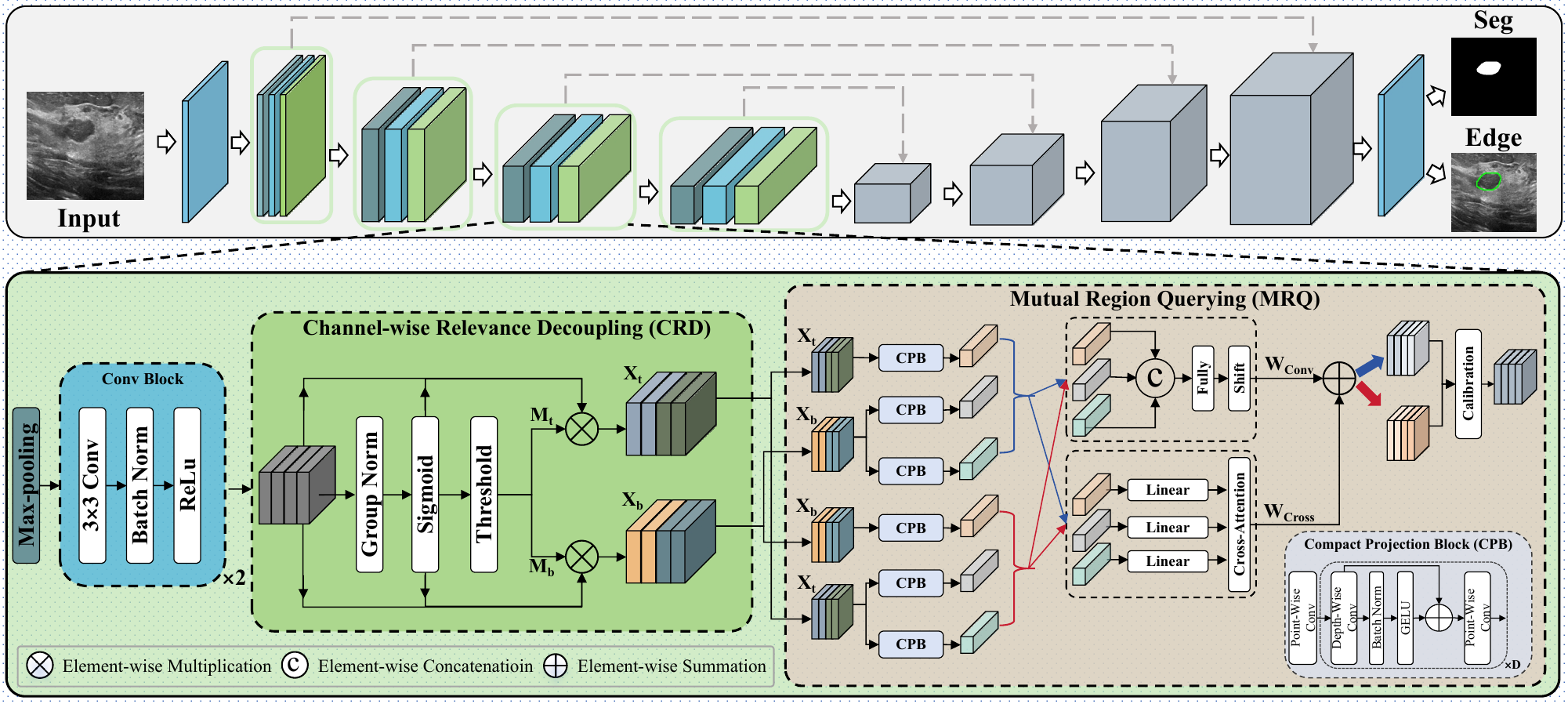}
   \caption{Illustration of our proposed Channel-Aware Region Extrication (CARE) pipeline. The top panel illustrates the entire network based on a hierarchical UNet backbone. The green frame at middle left highlight the spatial locations where the Channel-wise Relevance Decoupling (CRD) module is integrated into the encoders to purify multi-scale features. The bottom-right panel showcases the Mutual Region Querying (MRQ) module, in which the Calibration unit represents the Reciprocal Feature Calibration (RFC) Block. Detailed architectures of the Compact Projection Block (CPB) and a comprehensive symbol legend are provided at the bottom center.}
   \label{fig:model}
\end{figure*}

\subsection{Channel-Aware Region Extrication}

\subsubsection{Overview}
As shown in \figref{fig:model}, we propose the Channel-Aware Region Extrication (CARE) module to effectively address the limitations of conventional methods in extricating the target object from the entangled background. To achieve this object, we first proposed Channel-wise Relevance Decoupling (CRD) to explicitly distinguish target and background features by gated operations. Moreover, to further enhance feature purity, we propose Mutual Region Querying (MRQ) to filter essential representations from the initial separated features, ensuring that the informative components are retained. Consequently, this hierarchical split-selection process eliminates feature ambiguity, enabling the model to isolate target objects with high precision even in highly distracting regions.

Specifically, following the common paradigm of medical image segmentation, we adopt a UNet-based backbone that features a symmetric encoder-decoder structure augmented by long-range skip connections. This hierarchical design allows the model to effectively capture multi-scale spatial representations while facilitating gradient flow, thereby mitigating the risk of network degradation during the training process. Within this hierarchical framework, particular emphasis is placed on the encoder to ensure robust feature extraction from challenging ultrasound backgrounds. At each encoding stage, a max-pooling layer is first employed for spatial downsampling, facilitating higher-level semantic abstraction. After the standard convolution blocks, we further introduce the CARE module, which is specifically designed for context-target disentanglement. By integrating it immediately after feature extraction, the network can selectively enhance target-relevant information while suppressing redundant or misleading background features in ultrasound images. The refined features are subsequently propagated through the decoder, where they are fused with corresponding encoder features via skip connections. This process ensures that the final pixel-wise predictions are guided by both enriched semantic information and preserved spatial details, leading to more accurate boundary delineation even in the presence of highly similar background interference.

\subsubsection{Channel-wise Relevance Decoupling}
Based on our observations, ultrasound images exhibit distinct acoustic scattering properties. Specifically, the densely cellular tumor stroma possesses a significantly higher scattering cross-section compared to normal fibroglandular and adipose tissues. This physical discrepancy results in log-compressed backscatter intensity distributions that are approximately unimodal but characterized by distinct means. Driven by this inherent separability in the acoustic domain, we propose the Channel-wise Relevance Decoupling (CRD) as the inaugural stage of the CARE module.

As depicted in \figref{fig:model}, the CRD aims to provide an initial coarse-grained separation of the target from the entangled background regions.To provide an initial coarse-grained separation of target and background cues, CRD explicitly decomposes the input feature map into two complementary features. One emphasizes informative target-related responses, while the other aggregates less informative responses that are more likely associated with entangled background interference. The decomposition is performed as follows.

Given an input feature map $X \in \mathbb{R}^{B \times C \times H \times W}$, we first apply Group Normalization (GN) to stabilize the channel-wise feature distribution. This ensures that the subsequent relevance scores are computed on a standardized and comparable scale, whereby the normalized representation $\hat{X}$ is obtained as:
\begin{equation}
\hat{X} = \mathrm{GN}(X)
\end{equation}

Subsequently, a sigmoid activation is employed to generate a channel-wise relevance map $R{_m}$:
\begin{equation}
R{_m} = \sigma(\hat{X})
\end{equation}

This operation maps the normalized features into the range $(0, 1)$ and thus provides a soft estimation of channel importance. In this manner, the model is able to quantify the importance of each channel relative to the target object, providing a foundation for the subsequent explicit feature splitting.

To achieve explicit separation between target-related and background-related features, we further introduce a threshold-based binary routing mechanism. By applying an indicator function $\mathbb{I}(\cdot)$ with a pre-defined threshold $\tau$, we generate two complementary decoupling maps, $M_{t}$ and $M_{b}$, as follows:

\begin{equation}
M_{\text{t}} = \mathbb{I}(R{_m} > \tau), \quad
M_{\text{b}} = \mathbb{I}(R{_m} \le \tau)
\end{equation}

Specifically, the indicator function $\mathbb{I}(\cdot)$ assigns a value of 1 to elements that satisfy the threshold condition $\tau$, and 0 to all others. This hard-assignment ensures that each feature channel is strictly categorized into either the informative branch or the non-informative branch, thereby preventing feature leakage between the target and the background.

Finally, the original input feature $X$ is split into two complementary branches by computing the Hadamard product between the original features $X$, the relevance map $R_m$, and the corresponding decoupling maps $M_{t}$ and $M_{b}$, which ensures that $X_{t}$ contains the target-related components only, while the $X_{b}$ isolates the distracting background context by:
\begin{equation}
X_{t} = X \odot R{_m} \odot M_{\text{t}}, \quad
X_{b} = X \odot R{_m} \odot M_{\text{b}}
\end{equation}

In summary, by decoupling the entangled relevance in the channel dimension, the CRD module effectively eliminates feature ambiguity, establishing a robust basis for the subsequent spatial-aware refinement in the MRQ module. By leveraging a sigmoid-based gating mechanism, the CRD module can effectively identify these intensity-driven spectral differences within the high-dimensional channel space. The threshold-based gate mechanism ensures that $X_{t}$ (the base feature) primarily retains the high-scattering informative cues of the lesion, while $X_{b}$ (the query feature) isolates the confounding background context. This explicit partitioning provides the MRQ module with a decoupled dual-stream input, enabling precise feature filtering and spatial refinement.

\subsubsection{Mutual Region Querying}
\label{sub:fam}
Despite its effectiveness in channel-level decoupling, CRD remains a coarse-grained operation that can leave residual spatial noise. Another challenge is that subtle target cues might be misclassified into the non-informative branch $X_{b}$. To resolve these spatial ambiguities and recover potential information loss, a more granular refinement mechanism, the MRQ, as depicted in \figref{fig:model}, is proposed. Details of the MRQ are as follows.

The proposed MRQ comprises two symmetric interaction branches that perform reciprocal information distillation between $X_{t}$ and $X_{b}$. In the $b \rightarrow t$ branch, $X_{t}$ is projected as the Query to retrieve target-relevant information from $X_{b}$, which provides the Key and Value. Conversely, the $t \rightarrow b$ branch takes $X_{b}$ as the Query and uses $X_{t}$ as the Key and Value, thereby enabling bidirectional information exchange between the two feature representations.

Before the reciprocal interaction, $X_{t}$ and $X_{b}$ are transformed into their corresponding Query, Key, and Value embeddings through three Compact Projection Blocks (CPBs):
\begin{equation}
\footnotesize
\begin{aligned}
Q_t &= \mathrm{CPB}^{Q}_{t}(X_t), &
K_b &= \mathrm{CPB}^{K}_{b}(X_b), &
V_b &= \mathrm{CPB}^{V}_{b}(X_b).
\end{aligned}
\end{equation}
\begin{equation}
\footnotesize
\begin{aligned}
Q_b &= \mathrm{CPB}^{Q}_{b}(X_b), &
K_t &= \mathrm{CPB}^{K}_{t}(X_t), &
V_t &= \mathrm{CPB}^{V}_{t}(X_t).
\end{aligned}
\end{equation}

As illustrated in \figref{fig:model}, each CPB adopts a lightweight bottleneck architecture composed of D stacked depth-wise convolutions and GELU activations. By progressively encoding local spatial patterns, the CPB maps pixel-level features into compact, region-aware embeddings. These embeddings preserve fine-grained structural and semantic cues while reducing redundant responses, thus providing discriminative representations for the subsequent bidirectional querying process.

Upon completing the projection via CPB, the MRQ module executes two symmetric interaction branches, both integrating a convolution path and a cross-attention path to perform content-aware refinement. Specifically, in Branch 1, $X_{t}$ serves as the Query while $X_{b}$ as the Key and Value. The objective of this configuration is to leverage the background context to calibrate and sharpen the boundaries of the target regions. Formally, for the convolution path, we first aggregate the region embeddings:
\begin{equation}
Y^{(t)}_{\mathrm{conv}}=\mathrm{Shift}\big(\mathrm{FC}(\mathrm{Concat}(Q_t,K_b,V_b))\big),
\end{equation}

Simultaneously, the cross-attention path performs non-local spatial retrieval to aggregate complementary context, yielding the attention-based features \begin{equation}
Y^{(t)}_{\mathrm{cross}}=\mathrm{CA}(Q_t,K_b,V_b),
\end{equation}
where $\mathrm{CA}(\cdot)$ denotes the cross-attention mechanism. This path establishes long-range dependencies between the decoupled branches, enabling $X_{t}$ to re-align with high-confidence regional structures present in $X_{b}$, thereby ensuring spatial consistency during the feature refinement process.

The final output is determined through the integration of these two paths, achieved by the addition of $Y_{\mathrm{Conv}}^{(t)}$ and $Y_{\mathrm{Cross}}^{(t)}$, their respective contributions modulated by two learnable scalars (W$_{Conv}$ and W$_{Cross}$), as delineated in eq.~(\ref{eq: branch1_fuse}).
\begin{equation}
   \label{eq: branch1_fuse}
   Y^{(t)}=W_{\mathrm{conv}}\,Y^{(t)}_{\mathrm{conv}}+W_{\mathrm{cross}}\,Y^{(t)}_{\mathrm{cross}},
\end{equation}

In Branch 2, the roles are reversed: $X_{b}$ acts as the Query to interrogate $X_{t}$. This reciprocal process serves as an information recovery mechanism, proactively scanning the regions previously identified as "non-informative" to reinstate minute target cues that may have been erroneously suppressed during the initial decoupling:
\begin{equation}
\label{eq: branch2_concat}
Y^{(b)}_{\mathrm{conv}}=\mathrm{Shift}\big(\mathrm{FC}(\mathrm{Concat}(Q_b,K_t,V_t))\big),
\end{equation}

Similarly, the cross-attention path in Branch 2 enables the non-informative features to re-reference the target-aware context from $X_{t}$, facilitating the identification of residual target signals within the background branch:
\begin{equation}
\label{eq: branch2_att}
Y^{(b)}_{\mathrm{cross}}=\mathrm{CA}(Q_b,K_t,V_t),
\end{equation}
The integration of the convolution and cross-attention paths for Branch 2 is formulated as:
\begin{equation}
\label{eq: branch2_fuse}
Y^{(b)}=W_{\mathrm{conv}}\,Y^{(b)}_{\mathrm{conv}}+W_{\mathrm{cross}}\,Y^{(b)}_{\mathrm{cross}}.
\end{equation}
where $W_{\mathrm{Conv}}$ and $W_{\mathrm{Cross}}$ are also the learnable scalars used to maintain structural symmetry.

After obtaining $Y^{(t)}$ and $Y^{(b)}$, we introduce the Reciprocal Feature Calibration (RFC) block to reduce information redundancy and promote complementary information exchange between the two interaction branches. Specifically, each branch feature is evenly partitioned into two disjoint groups along the channel dimension:
\begin{equation}
Y^{(t)} = Y^{(t)}_1 \cup Y^{(t)}_2,\qquad
Y^{(b)} = Y^{(b)}_1 \cup Y^{(b)}_2,
\end{equation}
where $\cup$ denotes the channel-wise composition of two mutually exclusive feature groups. The resulting groups are then paired across the two branches and fused through element-wise addition. Finally, the calibrated feature is reconstructed as
\begin{equation}
Y_f=\operatorname{Concat}_{c}\left(
Y^{(t)}_1\oplus Y^{(b)}_2,\,
Y^{(t)}_2\oplus Y^{(b)}_1
\right),
\end{equation}
where $\oplus$ denotes element-wise addition and
$\operatorname{Concat}_{c}(\cdot)$ denotes channel-wise
concatenation. This crossed fusion allows complementary information
from one branch to compensate for potentially degraded responses in
the other branch while restoring the original feature dimensions.

In summary, the MRQ module transcends simple feature fusion by establishing a dual path information bridge. This mutual query-and-refine paradigm ensures that the decoupled features are not only purified but also spatially coherent, providing a highly discriminative representation for the subsequent operations.

\subsubsection{Training Objective}

To jointly optimize region segmentation and boundary-aware refinement, the overall training objective is defined as:
\begin{equation}
L_{\mathrm{total}} = L_{\mathrm{seg}} + \lambda_{1} L_{\mathrm{edge}} + \lambda_{2} L_{\mathrm{gate}} + \lambda_{3} L_{\mathrm{cons}},
\end{equation}
where $L_{\mathrm{seg}}$ is the primary segmentation loss, and $L_{\mathrm{edge}}$, $L_{\mathrm{gate}}$, and $L_{\mathrm{cons}}$ denote the auxiliary edge supervision, gate supervision, and branch consistency terms, respectively.
In our implementation, the weighting coefficients are empirically set to $\lambda_{1}=0.3$, $\lambda_{2}=0.1$, and $\lambda_{3}=0.1$ to balance the contributions of the auxiliary loss functions.

The segmentation loss combines Dice loss and binary cross-entropy (BCE) loss:
\begin{equation}
L_{\mathrm{seg}} = L_{\mathrm{Dice}}(\hat{M}, M) + L_{\mathrm{BCE}}(\hat{M}, M),
\end{equation}
where $\hat{M}$ and $M$ denote the predicted mask and the ground-truth mask, respectively.

To enhance boundary sensitivity, we further impose an auxiliary edge loss on the edge prediction branch:
\begin{equation}
L_{\mathrm{edge}} = L_{\mathrm{Dice}}(E, B) + L_{\mathrm{BCE}}(E, B),
\end{equation}
where $E$ denotes the edge prediction, and $B$ is the boundary band derived from the ground-truth mask:
\begin{equation}
B = \mathrm{Dilate}(M) - \mathrm{Erode}(M).
\end{equation}

For CARE-specific supervision, we define the gate supervision loss as:
\begin{equation}
L_{\mathrm{gate}} = L_{\mathrm{BCE}}(R_m, B_{\mathrm{ds}}),
\end{equation}
where $R_m$ is the relevance map generated by CRD, and $B_{\mathrm{ds}}$ denotes the downsampled boundary band.

In addition, we impose a branch consistency constraint on the two refined branches in MRQ:
\begin{equation}
L_{\mathrm{cons}}=
\sum_{j\in\{1,2\}}
\left\|
\frac{F_{1}^{j}}{\|F_{1}^{j}\|_{2}}
-
\frac{F_{2}^{j}}{\|F_{2}^{j}\|_{2}}
\right\|^{2}
\cdot B_{\mathrm{ds}},
\end{equation}
where $F_{1}$ and $F_{2}$ denote the two refined feature branches produced by MRQ.
The three auxiliary objectives regularize CARE from complementary perspectives. $L_{\mathrm{edge}}$ improves boundary sensitivity under weak contrast. Rather than using the full lesion mask, $L_{\mathrm{gate}}$ supervises the relevance map with a downsampled boundary band, guiding CRD to emphasize ambiguity-sensitive regions instead of the entire foreground. $L_{\mathrm{cons}}$ aligns the two refined branches around boundary-sensitive regions while preserving their complementary roles elsewhere. Together, these objectives stabilize representation extrication and improve target-context discrimination.

\begin{table*}[t]
\centering
\small
\setlength{\tabcolsep}{6.5pt}
\renewcommand{\arraystretch}{1.08}
\begin{threeparttable}
\caption{Comparison results on three datasets. The best results are shown in \textbf{bold}, and the second-best results are \underline{underlined}.}
\label{tab: Res_com_all}

\begin{tabular}{
l
S[table-format=2.2] S[table-format=2.2] S[table-format=2.2] S[table-format=2.2]
S[table-format=2.2] S[table-format=2.2] S[table-format=2.2] S[table-format=2.2]
S[table-format=2.2] S[table-format=2.2] S[table-format=2.2] S[table-format=2.2]
}
\toprule
\multirow{2}{*}{Methods}
& \multicolumn{4}{c}{BUSI (\%)}
& \multicolumn{4}{c}{BUSIS (\%)}
& \multicolumn{4}{c}{TN3K (\%)} \\
\cmidrule(lr){2-5}
\cmidrule(lr){6-9}
\cmidrule(lr){10-13}

& {DSC} & {mIoU} & {Precision} & {Recall}
& {DSC} & {mIoU} & {Precision} & {Recall}
& {DSC} & {mIoU} & {Precision} & {Recall} \\
\midrule

UNet~\cite{ronneberger2015u}
& 76.18 & 67.62 & 79.09 & 79.71
& 91.18 & 84.89 & 93.02 & 90.88
& 77.69 & 67.38 & 74.91 & 87.08 \\

ResUNet~\cite{zhang2018road}
& 77.27 & 68.45 & 79.20 & 80.36
& 91.26 & 85.09 & \underline{93.18} & 91.15
& 76.76 & 66.67 & 73.93 & 86.18 \\

AttUNet~\cite{oktay2018attention}
& 76.62 & 68.09 & 79.72 & 78.44
& 91.04 & 84.65 & 93.04 & 90.66
& 77.80 & 67.86 & 74.94 & 87.04 \\

TransUNet~\cite{chen2021transunet}
& 71.94 & 61.88 & 78.57 & 73.57
& 89.97 & 82.69 & 90.09 & 91.53
& 71.65 & 60.03 & 69.37 & 83.13 \\

UNeXt~\cite{valanarasu2022unext}
& 72.27 & 61.64 & 77.06 & 75.39
& 89.97 & 82.41 & 90.69 & 90.57
& 74.35 & 63.15 & 72.07 & 85.19 \\

CMUNet~\cite{tang2023cmu}
& 79.95 & 71.68 & \underline{84.13} & 81.45
& 91.43 & 85.28 & 92.86 & 91.58
& 79.86 & 70.15 & 78.35 & 85.19 \\

CMUNeXt~\cite{tang2024cmunext}
& 74.52 & 65.32 & 77.53 & 76.46
& 90.43 & 83.41 & 90.89 & 91.29
& 75.83 & 65.73 & 73.39 & 85.34 \\

EoMT~\cite{kerssies2025your}
& 81.44 & 72.47 & 81.53 & \underline{84.27}
& 91.93 & 85.63 & 92.28 & \underline{92.54}
& 81.58 & 71.57 & 79.95 & \underline{89.31} \\

MSAGHNet~\cite{zhu2025multi}
& 81.44 & 72.53 & 84.04 & 81.78
& 91.36 & 84.76 & 92.88 & 90.94
& 81.01 & 71.35 & 79.83 & 87.55 \\

PMRNet~\cite{Kang_2026_CVPR}
& \underline{81.74}
& \underline{73.54}
& 82.56
& 83.63
& \underline{91.96}
& \underline{85.96}
& 93.12
& 91.93
& \underline{81.93}
& \underline{72.42}
& \underline{80.77}
& 88.10 \\

\midrule

\textbf{CARE}
& \textbf{83.12}
& \textbf{75.14}
& \textbf{84.30}
& \textbf{84.39}
& \textbf{92.57}
& \textbf{86.79}
& \textbf{93.42}
& \textbf{92.56}
& \textbf{82.97}
& \textbf{74.00}
& \textbf{81.56}
& \textbf{89.42} \\

\bottomrule
\vspace{-0.6cm}
\end{tabular}
\end{threeparttable}
\end{table*}

\section{Experiments}
\label{sec:Experiments}
\subsection{Datasets and Evaluation Metrics}
\subsubsection{Datasets}The datasets utilized in our study are the Breast UltraSound Images (BUSI) dataset \cite{al2020dataset}, BUSIS dataset~\cite{zhang2022busis}, and TN3k dataset~\cite{gong2022thyroid, gong2021multi-task}. The BUSI dataset, a widely recognized benchmark dataset, consists of 780 breast ultrasound images, focusing on benign and malignant cases (647 images) randomly split for training (453 images), validation (65 images), and testing (129 images). The BUSIS dataset includes 562 images obtained from women aged 26 to 78 years, the images were de-identified and sourced from different hospitals, with random splits for training (394 images), validation (56 images), and testing (112 images). Finally, the TN3k dataset contributes 3493 thyroid ultrasound images, following the official data split with 2879 images for training and validation and 614 images for testing.

\subsubsection{Evaluation Metrics}
We employ four widely used metrics to quantitatively evaluate segmentation performance, including the Dice Similarity Coefficient (DSC), mean Intersection over Union (mIoU), Precision, and Recall.
Given the predicted binary mask $\hat{M}$ and the ground-truth mask $M$, the overlap-based metrics are defined as follows:
\begin{equation}
\mathrm{DSC} = \frac{2|\hat{M} \cap M|}{|\hat{M}| + |M|}, \qquad
\mathrm{mIoU} = \frac{|\hat{M} \cap M|}{|\hat{M} \cup M|}.
\end{equation}

In addition, we report Precision and Recall to evaluate the pixel-level correctness of positive predictions and the completeness of lesion recovery, respectively:
\begin{equation}
\mathrm{Precision} = \frac{TP}{TP + FP}, \qquad
\mathrm{Recall} = \frac{TP}{TP + FN},
\end{equation}
where $TP$, $FP$, and $FN$ denote the numbers of true positive, false positive, and false negative pixels, respectively.

\subsection{Implementation Details}
\textbf{Training details.} All of our experiments were conducted using PyTorch 1.7.0. We trained all models on a single NVIDIA RTX A6000 GPU with 48GB of memory. The Adam optimizer was employed with a learning rate annealing factor of 0.2, initializing at 1e-4, and a weight decay of 5e-4.

\textbf{Data Preprocessing.} Due to the limitation of the GPU memory, we use a batch size of 8 and train each model for 100 epochs. Besides, following the setting of~\cite{tang2023cmu} we resize all the images as $256 \times 256 \times 3$. Moreover, considering that insufficient data will induce overfitting, we apply data augmentation on the training set to alleviate this phenomenon. Specifically, we leverage random rotation, flip, elastic transform, and light transforms on the images with a probability of 0.5. Note that no operations were performed on the validation and test sets.

\begin{table}[t]
\centering
\caption{Ablation study results on BUSI, BUSIS, and TN3K datasets.}
\label{table: ablation}
\small
\setlength{\tabcolsep}{3pt}
\begin{tabular}{l|ccc|cccc}
\hline
\multirow{2}{*}{\bf Datasets} & \multicolumn{3}{c|}{\bf Components} & \multicolumn{4}{c}{\bf Metrics (\%)} \\ \cline{2-8}
 & Baseline & CRD & MRQ & DSC & mIoU & Precision & Recall \\ \hline
\multirow{3}{*}{\bf BUSI}
& \cmark & \xmark & \xmark & 76.18 & 67.62 & 79.09 & 79.71 \\
& \cmark & \cmark & \xmark & 80.52 & 71.81 & 83.16 & 81.50 \\
& \cmark & \cmark & \cmark & \bf 83.12 & \bf 75.14 & \bf 84.30 & \bf 84.39 \\ \hline
\multirow{3}{*}{\bf BUSIS}
& \cmark & \xmark & \xmark & 91.18 & 84.89 & 93.02 & 90.88 \\
& \cmark & \cmark & \xmark & 91.86 & 85.84 & 92.55 & 92.33 \\
& \cmark & \cmark & \cmark & \bf 92.57 & \bf 86.79 & \bf 93.42 & \bf 92.56 \\ \hline
\multirow{3}{*}{\bf TN3K}
& \cmark & \xmark & \xmark & 77.69 & 67.38 & 74.91 & 87.08 \\
& \cmark & \cmark & \xmark & 79.49 & 69.81 & 77.27 & 87.29 \\
& \cmark & \cmark & \cmark & \bf 82.97 & \bf 74.00 & \bf 81.56 & \bf 89.42 \\ \hline
\end{tabular}
\end{table}

\begin{figure*}[ht]
   \centering
   \includegraphics[width=0.95\textwidth]{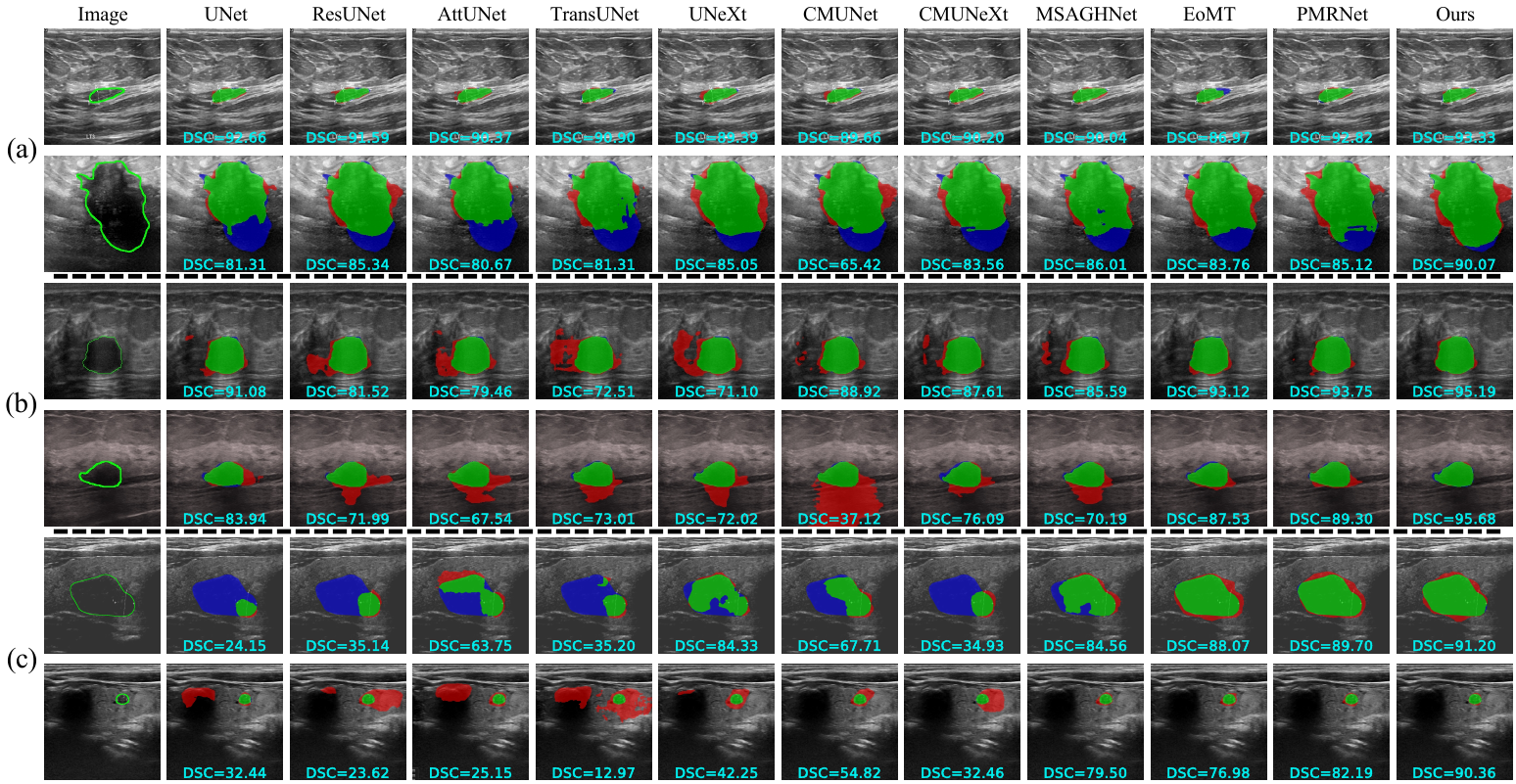}
     \vspace{-0.4cm}
   \caption{Visualized results on BUSI, BUSIS, and TN3K datasets. From left to right are the input image, results of UNet, ResUNet, AttUNet, TransUNet, UNeXt, CMUNet, CMUNeXt, MSAGHNet, EoMT, PMRNet, and our proposed method, respectively. (a) represents the segmentation results of benign and malignant nodules from the BUSI dataset; (b) represents segmentation results on the BUSIS dataset; and (c) represents results on the TN3K dataset. The \textcolor{fillgreen}{green}, \textcolor{fillblue}{blue} and \textcolor{fillred}{red} regions indicate the correct, missed and wrong predictions compared to the ground-truth mask.}
   \label{fig: res_all}
     \vspace{-0.6cm}
\end{figure*}

\subsection{Results}

We compare \textbf{CARE} with ten representative segmentation models, including UNet, ResUNet, AttUNet, TransUNet, UNeXt, CMUNet, CMUNeXt, MSAGHNet, EoMT and PMRNet.
The quantitative results on BUSI, BUSIS, and TN3K are reported in Table~\ref{tab: Res_com_all}.

Overall, CARE achieves the best performance on all three datasets across the four evaluation metrics.
Compared with the strongest competing method, CARE improves DSC/mIoU by 1.38\%/1.60\% on BUSI, 0.61\%/0.83\% on BUSIS, and 1.04\%/1.58\% on TN3K, respectively.
These consistent gains across datasets indicate that CARE is effective under different ultrasound imaging conditions and lesion distributions.
Notably, the improvements are not limited to overlap-based metrics. CARE also achieves the best Precision and Recall on all three datasets, which suggests that the performance gains are not obtained by simply predicting larger lesion regions.
Instead, the simultaneous improvement in Precision and Recall indicates that CARE is able to better preserve true lesion responses while suppressing visually similar distractors.
This observation is consistent with our motivation that ultrasound segmentation is limited by insufficient target-context discrimination rather than coarse localization alone.

In addition, the improvements are particularly notable on BUSI and TN3K, where CARE surpasses the previous best method by 1.60\% and 1.58\% in mIoU, respectively.
We attribute this to the stronger visual ambiguity and background interference in these datasets, where explicit representation extrication becomes more beneficial.
The relatively smaller but still consistent gain on BUSIS further suggests that CARE remains effective even when the baseline performance is already satisfactory.

\figref{fig: res_all} provides qualitative comparisons.
Compared with competing methods, CARE produces segmentation masks that are more complete, more compact, and better aligned with the ground truth.
In challenging cases with small nodules, weak boundaries, or strong background resemblance, previous methods tend to over-segment distracting regions, miss subtle lesion parts, or generate irregular boundaries.
By contrast, CARE more reliably captures the full lesion extent while reducing false positive activations in surrounding tissues.
This advantage is also evident in cases containing multiple nodules, where CARE can simultaneously preserve target completeness and suppress unrelated background interference.
These visual results further support that CARE improves lesion delineation by enhancing target-context discrimination under severe visual ambiguity.

\begin{figure}[ht]
   \centering
   \includegraphics[width=\linewidth]{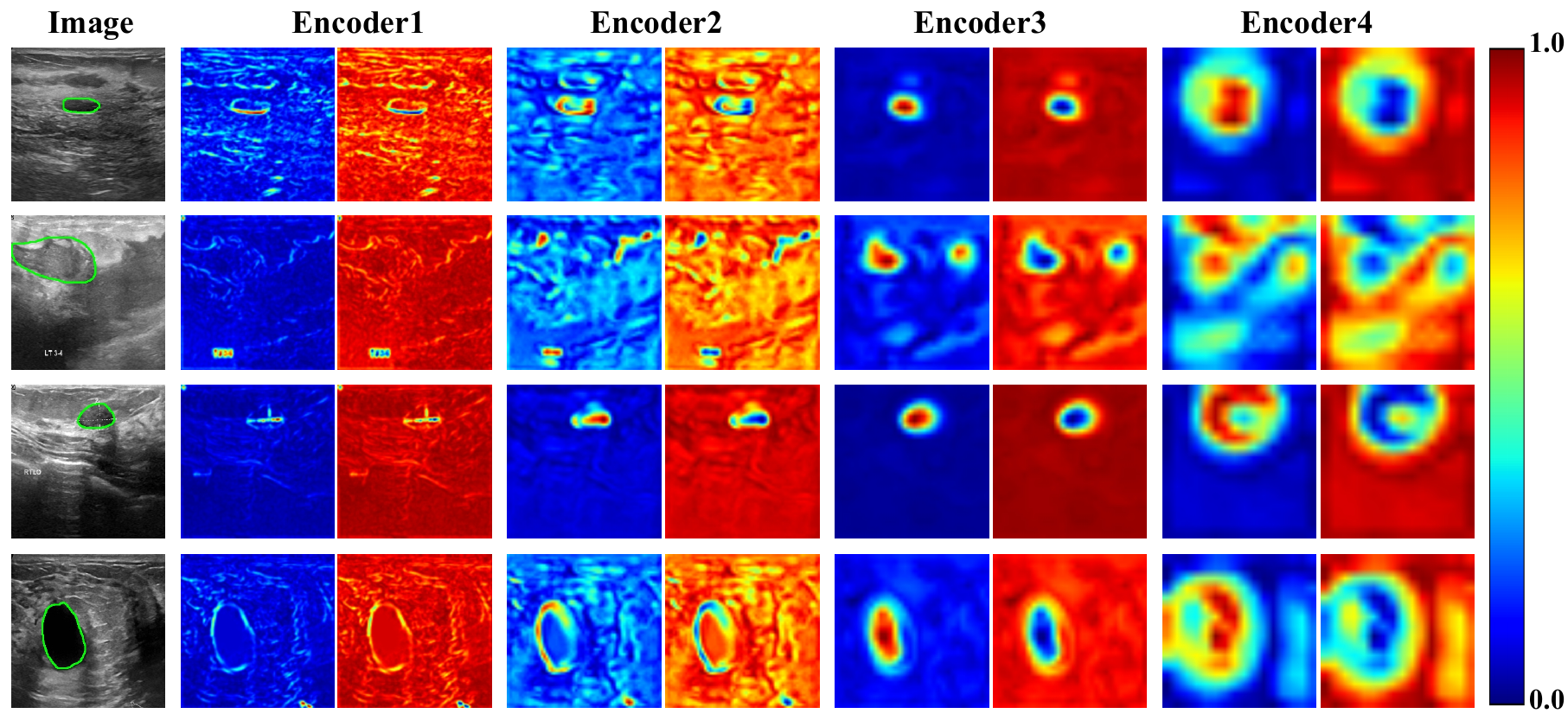}
     \vspace{-0.8cm}
\caption{Visualization of channel-averaged difference maps. For each encoder stage, the feature discrepancy between the decoupled branches is illustrated. The left column shows $X_t - X_b$ and the right column displays $X_b - X_t$. Prior to subtraction, channel-wise averaging is performed on each feature map.}
   \label{fig: ablation}
     \vspace{-0.8cm}
\end{figure}

\begin{figure}[ht]
   \centering
   \includegraphics[width=\linewidth]{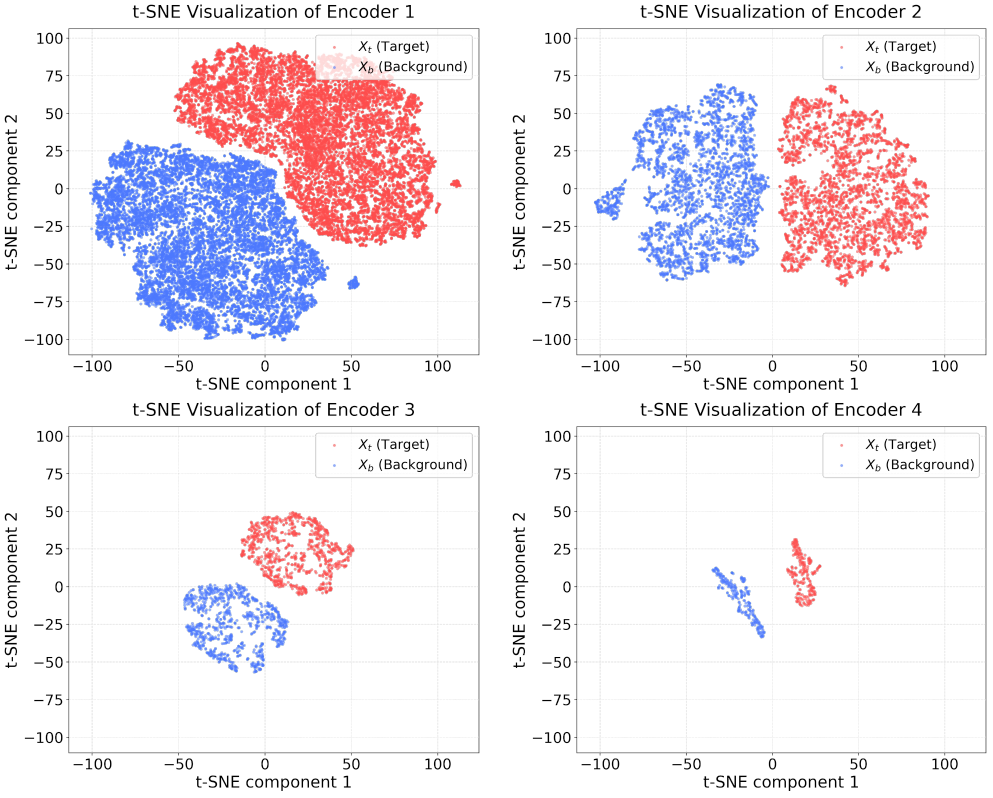}
     \vspace{-0.6cm}
    \caption{t-SNE visualization of the decoupled features from each encoder stage. Red points correspond to the lesion-relevant branch $X_t$, while blue points correspond to the interference-dominant branch $X_b$. The two branches form distinguishable clusters with limited overlap, and their separation becomes clearer in deeper encoder stages. This trend suggests that CRD progressively organizes lesion-relevant and background-interference cues into more separable representations during feature encoding.}
   \label{fig: tsne}
     \vspace{-0.6cm}
\end{figure}

\subsection{Ablation Study}

\subsubsection{Quantitative and Qualitative Analysis}
The ablation results are summarized in Table~\ref{table: ablation}.
We evaluate the contribution of the two key components in CARE, namely the initial channel-wise decoupling and the subsequent reciprocal refinement.
Starting from the baseline model, introducing CARE without MRQ already yields clear improvements on all three datasets.
Specifically, on BUSI, the DSC and mIoU increase by 4.34\% and 4.19\%, respectively, while on BUSIS the corresponding gains are 0.68\% and 0.95\%.
On TN3K, the same variant improves DSC and mIoU by 1.80\% and 2.46\%.
These results indicate that the initial channel-aware decoupling is already effective in reducing target-context ambiguity.

When MRQ is further introduced, the performance is consistently improved again across all datasets.
Compared with the variant without MRQ, the full model brings additional gains of 2.60\%/3.33\% in DSC/mIoU on BUSI and 0.71\%/0.95\% on BUSIS.
On TN3K, the improvement becomes more pronounced, reaching 3.48\% in DSC and 4.19\% in mIoU.
This suggests that reciprocal refinement provides complementary benefits beyond the initial decoupling, especially in more challenging cases with stronger background interference.
Overall, the ablation study confirms that both CRD and MRQ are necessary, and that the best performance is achieved when channel-aware separation and reciprocal refinement are used together.

To further examine the effect of CRD, we visualize the differential activation patterns of the two branches in \figref{fig: ablation}.
The map $X_t - X_b$ produces stronger responses around lesion regions, whereas $X_b - X_t$ highlights distracting background context.
The contrasting activation patterns suggest that the two branches emphasize different types of information after decoupling, which is consistent with the design objective of separating lesion-relevant and interference-dominant responses.
We also visualize the feature distributions of the two branches at all four encoder stages using t-SNE in \figref{fig: tsne}.
Across all stages, the features from $X_t$ and $X_b$ form distinguishable clusters with limited overlap, indicating that CRD produces separable representations in feature space.
Moreover, the separation becomes clearer in deeper encoder stages, suggesting that the decoupling effect is progressively reinforced along the hierarchical encoding process.
These observations provide additional evidence that CARE improves ultrasound segmentation by explicitly organizing complementary lesion and background cues in the learned representation.

\begin{table}[h]
    \centering
    \footnotesize
    \captionsetup{font=footnotesize,skip=2pt}
    \caption{Fine-grained ablation study of CARE on BUSI.}
    \label{tab:comprehensive_ablation_busi}
    \setlength{\tabcolsep}{4pt}

    \begin{tabular}{lcc}
        \toprule
        Methods & DSC (\%) & mIoU (\%) \\
        \midrule
        CARE \emph{w/o} CPB & 82.12 & 73.91 \\
        CARE \emph{w/o} RFC & 81.86 & 73.52 \\
        \midrule
        CARE \emph{w/o} Gating & 81.68 & 73.57 \\
        CARE \emph{w/} Soft Gating & 81.84 & 73.49 \\
        \midrule
        CARE \emph{w/o} Cross-Attention Path & 82.06 & 73.32 \\
        CARE \emph{w/o} Convolution Path & 82.42 & 73.70 \\
        \midrule
        CARE \emph{w/o} Edge Loss & 82.17 & 73.86 \\
        CARE \emph{w/o} Gate Loss & 82.63 & 74.41 \\
        CARE \emph{w/o} Consistency Loss & 82.76 & 74.67 \\
        \midrule
        \textbf{Full CARE} & \textbf{83.12} & \textbf{75.14} \\
        \bottomrule
    \end{tabular}
\end{table}

\begin{table}[t]
\centering
\vspace{-0.6cm}
\caption{Quantitative comparison of our model performance under different threshold ($\tau$) settings across three datasets. The best results are highlighted in bold.}
\label{table:threshold}
\small \setlength{\tabcolsep}{4pt}
\resizebox{\columnwidth}{!}{
    \begin{tabular}{l|c|cccc}
    \toprule
    {\bf Datasets} & {\bf $\tau$} & {\bf DSC (\%)} & {\bf mIoU (\%)} & {\bf Precision (\%)} & {\bf Recall (\%)} \\
    \midrule
    \multirow{3}{*}{BUSI}  & 0.4 & 82.71 & 74.65  & 83.62 & \textbf{84.95}  \\
                           & 0.5 & \textbf{83.12} & \textbf{75.14} & 84.30 & 84.39 \\
                           & 0.6 & 82.84  & 74.61 & \textbf{84.64}  & 84.35 \\
    \hline
    \multirow{3}{*}{BUSIS} & 0.4 & 92.10  & 86.17 & 92.42  & \textbf{93.00} \\
                           & 0.5 & \textbf{92.57} & \textbf{86.79} & 93.42 & 92.56 \\
                           & 0.6 & 92.28  & 86.35  & \textbf{93.57}  & 91.95 \\
    \hline
    \multirow{3}{*}{TN3K}  & 0.4 & 82.72  & 73.63  & 81.37  & 89.14 \\
                           & 0.5 & \textbf{82.97} & \textbf{74.00} & \textbf{81.56} & 89.42 \\
                           & 0.6 &82.70  &73.65  &81.04  &\textbf{89.51} \\
    \bottomrule
    \end{tabular}
}
\end{table}

\subsubsection{Fine-grained Component Analysis}

Beyond the module-level analysis in Table~\ref{table: ablation},
we further conduct fine-grained ablation experiments on BUSI to
examine the individual contributions of the internal components,
routing strategies, interaction paths, and auxiliary objectives in
CARE. The corresponding results are reported in
Table~\ref{tab:comprehensive_ablation_busi}.

\noindent
Removing CPB and RFC reduces DSC/mIoU by 1.00\%/1.23\% and 1.26\%/1.62\%, respectively, confirming the benefits of region-aware projection and cross-branch calibration. Removing gating causes drops of 1.44\%/1.57\%, while replacing hard gating with soft gating decreases performance by 1.28\%/1.65\%, demonstrating that binary routing more effectively separates lesion-relevant and interference-dominant responses. Removing the cross-attention and convolution paths in MRQ reduces DSC/mIoU by 1.06\%/1.82\% and 0.70\%/1.44\%, respectively, validating their complementary roles in non-local interaction and local-detail preservation. Moreover, removing the edge, gate, and consistency losses decreases DSC by 0.95\%, 0.49\%, and 0.36\%, respectively, with edge supervision contributing the most. Overall, these results verify the contributions of all CARE components.

\subsubsection{Threshold Sensitivity Analysis}

To evaluate the robustness of CARE to the routing threshold in CRD, we vary $\tau$ from 0.4 to 0.6 and report the results in Table~\ref{table:threshold}.
Overall, the performance remains stable across different settings, indicating that CARE is not overly sensitive to the threshold choice.
Among the tested values, $\tau=0.5$ achieves the best overall balance across evaluation metrics, thereby validating it as the optimal choice adopted for all experiments.

\subsubsection{Complexity Analysis}

\begin{table}[t]
\centering
\setlength{\tabcolsep}{5pt}
\renewcommand{\arraystretch}{1.08}
\vspace{-0.4cm}
\caption{Complexity comparison of representative segmentation models. FLOPs are computed on an input size of $256\times256\times3$. Lower is better for both metrics.}
\label{tab:efficiency}
\begin{tabular}{lcc}
\toprule
\textbf{Methods} & \textbf{FLOPs$\downarrow$ (G)} & \textbf{Params$\downarrow$ (M)} \\
\midrule
UNet        & 40.16 & 17.26 \\
AttUNet     & 66.68 & 34.88 \\
TransUNet   & 33.69 & 67.08 \\
CMUNet      & 91.33 & 49.93 \\
CMUNeXt     & 7.41  & \textbf{3.15} \\
MSAGHNet    & \textbf{2.16} & 23.47 \\
\midrule
\textbf{CARE} & 77.04 & 40.58 \\
\bottomrule
\vspace{-0.8cm}
\end{tabular}
\end{table}

To further assess the computational cost of CARE, we compare its FLOPs and parameter count with several representative segmentation models in Table~\ref{tab:efficiency}.
CARE is not the lightest model, since the reciprocal interaction in MRQ and the repeated refinement across encoder stages introduce additional computation beyond plain encoder--decoder baselines.
Nevertheless, CARE provides a favorable accuracy--complexity trade-off among strong competing methods.
For example, CARE uses fewer parameters than TransUNet (40.58M vs.\ 67.08M), and is also lighter than CMUNet in both FLOPs and parameter count (77.04G/40.58M vs.\ 91.33G/49.93M), while achieving the best segmentation performance on BUSI, BUSIS, and TN3K.

\section{Conclusion}
\label{sec:Conclusion}

In this paper, we present Channel-Aware Region Extrication (CARE) for ultrasound image segmentation.
Our work is motivated by the observation that ultrasound segmentation is fundamentally challenged by \emph{target-context entanglement}, where lesion cues are easily mixed with visually similar surrounding responses.
To address this issue, CARE improves segmentation by organizing lesion-relevant and interference-dominant cues in the learned representation, rather than relying solely on stronger feature extraction or context aggregation.
Within this framework, CRD provides an initial channel-aware separation of complementary responses, while MRQ further refines them through reciprocal interaction.
Extensive experiments on BUSI, BUSIS, and TN3K demonstrate the effectiveness and robustness of CARE in addressing target-context ambiguity.

\end{document}